Article

# A new Reinforcement Learning framework to discover natural flavor molecules

Luana P. Queiroz*, Carine M. Rebello[1], Erbet A. Costa[1], Vinícius V. Santana[2], Bruno C. L. Rodrigues[2], Alírio E. Rodrigues[2], Ana M. Ribeiro[2] and Idelfonso B. R. Nogueira[3]

*1*   *Departamento de Engenharia Química, Escola Politécnica (Polytechnic Institute), Universidade Federal da Bahia, Salvador 40210-630, Brazil;*

*2*   *Laboratory of Separation and Reaction Engineering, Associate Laboratory LSRE/LCM, Department of Chemical Engineering, Faculty of Engineering, University of Porto, Rua Dr. Roberto Frias, 4200-465 Porto, Portugal;*

*3*   *Chemical Engineering Department, Norwegian University of Science and Technology, Sem Sælandsvei 4, Kjemiblokk 5, Trondheim, Norway;*

*\* Correspondance: up201700139@edu.fe.up.pt*

**Abstract**

The flavor is the focal point in the flavor industry, which follows social tendencies and behaviors. The research and development of new flavoring agents and molecules are essential in this field. On the other hand, the development of natural flavors plays a critical role in modern society. In light of this, the present work proposes a novel framework based on Scientific Machine Learning to undertake an emerging problem in flavor engineering and industry. Therefore, this work brings an innovative methodology to design new natural flavor molecules. The molecules are evaluated regarding the synthetic accessibility, the number of atoms, and the likeness to a natural or pseudo-natural product.

# 1 Introduction

The sensation of a flavor is defined by the Encyclopaedia Britannica as the attribute of a substance perceived within the mouth produced by the senses of smell, taste, and touch [1]. From an anatomy point of view, this sensation is distinguished by the taste buds in the oral cavity, which is situated in the pharynx, palate, larynx, and tongue. An adult has approximately 10000 taste buds. The flavor response in the taste buds happens by identifying chemicals present in food and beverages, for example, and their translation into nerve signals. Understanding the flavor biological process and response is pivotal in developing the food, beverage, and flavor industries [2].

The flavor plays an essential role in several products found in nowadays markets. Furthermore, new flavors are required in manufacturers for product development and innovation. In this context, researching new flavors and developing innovative production technologies are constantly sought. On the other hand, society is observing an increasing consumption of processed products and improved fast-food products. These products are composed of several food additives and flavoring agents, alongside. More recently, expanding awareness of the importance of a healthy lifestyle has provoked a rise in the search for foods labeled with a natural flavoring agent. This movement has motivated the growth of flavors' R&D activities and investments in new flavor-based products [3]. In 2020, the flavors and fragrances' market size from food and beverage products were appraised at € 26.53 billion, and the prospect is that by 2026 it will reach € 35.52 billion [4].

Although synthetic flavors are profitable and largely applied in food products, the heaviest impact on the market share is from natural flavors. This is a consequence of the intensified search for a healthy lifestyle and the increasing awareness of the hazardous effects of some synthetic flavors. The increase of this new lifestyle has also impacted the

Beverage sector, which is continually growing and significantly influences the market. Overall, the largest end-use industries of flavors are beverages, bakery, savory and snacks, dairy and frozen products, confectionery, and pet food [5]. Such sectors are forecast to grow through increasing application and development of new products based on natural flavor [3]. However, other industries are also investing in flavoring their products to satisfy the consumers or make their products more appealing, such as producing flavored toothpaste.

According to the Regulation (EC) No 1334/2008 of the European Parliament and of the Council of 16 December 2008 on flavorings and certain food ingredients with flavoring properties for use in and on foods and amending Council Regulation (EEC) No 1601/91, Regulations (EC) No 2232/96 and (EC) No 110/2008 and Directive 2000/13/EC [6]:

• There is no distinction between nature-identical and artificial flavoring substances; both are referred to as "Flavoring substances";

• The labeling of product's ingredients as "natural flavoring substances" can only be applied exclusively for natural flavoring substances;

• Flavoring substances are defined as chemical substances obtained from chemical synthesis or isolated through chemical processes and natural flavoring substances;

• "Natural flavoring substance" is a flavoring substance produced by appropriate physical, enzymatic, or microbiological processes from a material of vegetable, animal, or microbiological origin.

The replication of natural flavor chemicals by synthetic molecules empowers the design of more stable, purer, potent, and cost-effective synthetic flavors. The discovery and combining of the molecules to express the multisensorial complexity as a nerve signal inherent to flavors is a trial-and-error process [8]. Additionally, the law and regulations must be taken into consideration when developing flavor-based products, especially concerning the environmental impact that the synthetic chemicals process can cause and the possible pathological state [10]. On that account, flavor engineering development is high-costly and time-consuming. In this way, the implementation of innovative technologies can be resourceful in reducing overheads and increasing efficiency. Consequently, Scientific Machine Learning (SciML) put forward a new approach to this industry.

Machine Learning (ML) tools are driving advances in the scientific field, especially the Scientific Machine Learning adaptation, which is an emergent area in the addressing of specific domain challenges. SciML is a resource-saving and efficient method in general data mining, modular design, image processing, bioinformatics, game playing, and computational chemistry [11]. The implementation of machine learning in chemistry engineering is optimistic in designing, synthesizing, and generating molecules and materials [12]. Particularly, a modest emerging but promising tendency in implementing SciML tools is perceived in flavor engineering.

Leong et al. (2021) [7] applied machine learning to analyze and quantify the flavor in a matrix. The authors developed a Surface-Enhanced Raman Scattering (SERS) taster that achieved outstanding accuracy in quantifying wine flavors using four receptors of the spectroscopic profiles. The impressive results were obtained to respond to the struggles of the chemical analysis of flavor compounds. It made it possible to analyze the detection in a rapid and sensitive procedure. Bi et al. (2020) [8] also applied ML to improve other aspects of flavor engineering. In the mentioned case, an approach was developed to predict olfactometry results from Gas Chromatography-Mass Spectrometry/Olfactometry (GC-MS/O). The olfactometry results can be achieved in 30 s with the proposed technology. However, the machine learning model presented an overfitting problem. The authors concluded that the technology could be improved even though it already has potential for olfactometry evaluation.

Although SciML applications can be found in the flavor engineering field, the whole potential of the technique is not yet deployed. The implementation of SciML in identifying and designing new flavoring molecules is not yet explored in the field. Consequently, the SciML is a simple and reliable way to approach these challenges and produce new chemical molecules with a specific flavor that can be synthesized. These goals were tackled in the context of this work.

Hence, this work proposes a novel framework combining generative neural network models and reinforcement learning to develop new flavors and flavor-based products. This allows the possibility of generating new flavor molecules that can be synthesized and applied in the industry to provide the products that can be considered natural. Thus, addressing a critical challenge found in the modern flavor industry.

**2 Methodology and results**

Reinforcement Learning (RL) was introduced by Richard Sutton and Andrew Barto in 1984 based on behavioral psychology and is the third machine learning paradigm, along with supervised and unsupervised learning. RL's application consists of an agent interacting and learning from an environment. Based on the experience amassed, the agent optimizes the defined objectives as cumulative rewards that must be maximized. The agent does not need to have all the information regarding the environment. It only requires the ability to interact with the environment and learn from it over time. This agent-environment relation is based on value function, reward, policy, and model. Recently, the efficiency of RL in solving sequential decision-making problems has increased its popularity [9,10].

The sequential function of the RL model starts with the agent receiving an initial state based on the observation of the environment. At a time, the agent must take action. Each action leads to a consequence that can be a reward, state change, and observation. The information gathered from its actions' results makes it possible for the agent to learn its purpose over time [10].

Artificial Intelligence (AI) development for video games has been a largely studied field aiming to achieve human-level performances when testing video games [11–13]. The complex interactions between the agent and the virtual environment involve a decision-making problem. In this instance, the RL model presents itself as a feasible solution. Solving game challenges with RL has shown excellent results, alongside its combination with deep learning to improve the generalization and scalability of the model [13].

The principle of this work is the development of new molecules with specific sought-after properties and flavors. The generative model designs new molecules through randomly chosen actions, even though they are associated with a probability of choice. The deep generative model is a resourceful technique to recognize patterns of likelihood between samples, which through the structure with numerous hidden layers of neural networks. Implementing a Reinforcement Learning algorithm alongside the deep neural network model of the generative model, maximizes the flavor-likeness of the molecule while carrying other desired properties [14]. Hence, a deep reinforcement learning process is proposed.

The methodology employed in this section is illustrated in Figure 1.

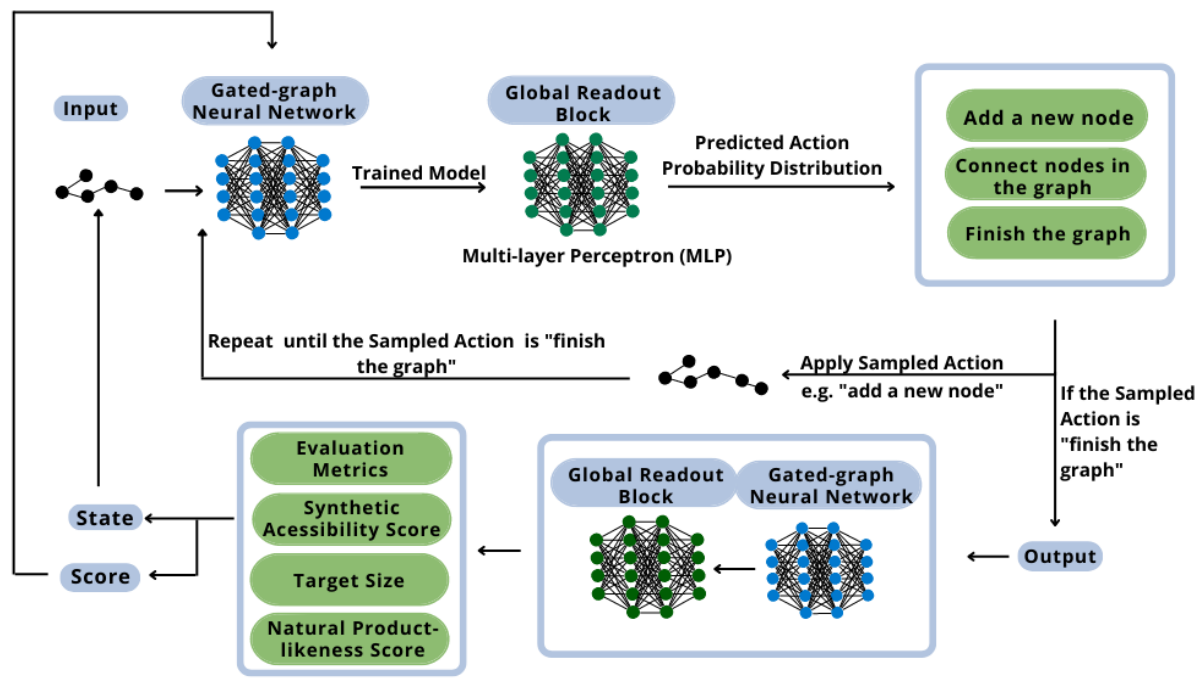

*Figure 1 Deep Reinforcement Learning model's methodology scheme.*

The first step is the sampling of database. In the development of this work a web scraper code developed extracted the database from the website FlavorDB [15]. For this purpose, 921 valid molecules were retrieved along with chemical properties information and the flavor descriptors associated with the corresponding molecules. In sum, 417 flavor descriptors were identified, in which the descriptors are correlated to each other.

Afterwards, each molecule of the database, one a one, are given as input to the deep generative model in the SMILES form. The main focus of this work is the implementation of reinforcement learning associated with the generative model.

First, the molecule is preprocessed so the model learns how to construct and deconstruct it following the canonical deconstruction path. Then, the obtained information follows the network path through the Gated Graph Neural Network (GGNN), introduced by Li et al. (2016) [16]. The resultant molecules are evaluated through evaluation metrics in intervals of epochs. The focus of this work is the implementation of reinforcement learning associated with the generative model, so the structure and methodology of the deep generative model will not be described.

After the generative model is trained, the trained models are saved. The best-trained model in the generative neural network is the one used to train the proposed deep reinforcement framework. The database utilized in the training of the generative model is once again given as an input. The system of GGNN and global readout block is kept in the same manner of the prior architecture. However, the agent impacts the way the decisions are made to build a molecule.

The possible actions are the same, being add a new node, connect nodes in the graph or finish the graph; simultaneously, the agent influences the choice of action and can influence the choice of atom type used in the valid molecule's design [17]. The environment states are the set of all the chemical structures designed during the generation process. In other words, the environment is the generative model, and the states are the output of the generation process for the whole database. Deep reinforcement learning takes the batch of generated molecules as input for the agent, not each molecule as the generative model does. It is essential to specify that, regardless of the reinforcement's input, the generative model as the environment receives each molecule at a time.

To achieve better results, maximize the rewards, the weights in the CNN policy are updated accordingly to the performance of the model. The CNN's weights also impact the results of the loss function. In these circumstances, the agent influence is in the weights in the calculation behind the generative model [17].

In a more straightforward perspective, compared to a video game, the environment is the deep generative model, the main character is the loss function's weight, and the state is the output of the deep generative model.

The generated molecules are analyzed and receive a reward from the agent. These molecules are, once again, deconstructed and reconstructed, but this time, in the agent's GGNN model. The mentioned process takes place to calculate the Action Probability Distribution (APD) and its influence on the loss function, in the same way as performed in the generative part. Similar to the previous section, the APD is calculated through a SoftMax function [18]. The result obtained by the agent for the current RL batch is compared to the one obtained in the previous batch.

The new action determined is applied to all the generated molecules aiming to estimate the success of this action. In other words, it predicts the number of molecules that will be fully terminated and valid when the generative model is reset with the updated weights. In this instance, the action that performs better is the one with the heaviest weight in the loss function. The metrics applied in analyzing a molecule's validation are the same as in the previous section, namely: section 2.2 .

The designed molecules given as outputs of the generative model, which are fully finished and valid, are separated from the rest of the batch to a group of finished molecules. These molecules are given as an output of the agent for the corresponding batch. One relevant aspect of this segmentation is that the RL's model considers the Uniformity-Completeness Jensen–Shannon Divergence (UC-JSD) metric as the generation metric. For this instance, the generation's success is also evaluated based on this specific metric.

Another critical part of the "traditional" reinforcement learning structure, used in games, is the reward. The deep reinforcement learning proposed here also makes use of rewards. The definition of the rewards is essential for guiding the agent's influence on the system, as well as to enforce the system to follow the properties of interest in designing molecules. In this work, the final score comprises the Synthetic Accessibility Score (SAScore) rewards, the target size, and the Natural Product-likeness Score (NPScore).

Ertl et al. (2009) [19] proposed the Synthetic Accessibility Score. The score is calculated considering the contribution of the molecule complexity and the fragmented contributions. It is based on previous studies of a million known chemicals that were already synthesized. The molecule complexity score notices the possible presence of non-standard structural features. The final SAScore for each molecule ranges from one to ten. An SAScore of one means that the molecule is easily synthesized, and ten means that the molecule is very difficult to synthesize. The reward was defined so that the SAScore value is inversely proportional to the reward given.

The Natural Product-likeness Score was introduced by Ertl et al. (2008) [20]. The score is a Bayesian measure to quantify the similarity of a compound with a natural compound based on the analysis of its structural fragment. The reference source for the natural compounds is the CRC Dictionary of Natural Products (DNP). The score is ranged between -5 and 5; the higher the score more similar to a natural product the molecule analyzed is. Around the value zero are the pseudo-natural products [21]. The term was proposed by Karageorgis et al. (2020) [22] and is applied to small molecules derived through the combination of natural products' fragments.

The target size is defined as follows: if the number of nodes in the molecule's graph is between zero and the predefined maximum, the reward is one, otherwise is zero. The final score is the sum of the rewards for each RL batch, in which each contribution in the score has equal weight. However, the final score calculation does not consider the contribution of molecules that are not unique, invalid, or fully finished.

The agent's policy applied in this work has a memory saving of the performance of the previous agents until the current state and is updated at every sampled learning step. It is based on the reward shaping mechanism introduced by Buhmann et al. (2011) [23]. This remembering makes it possible for the agent to know the actions that led to the best-designed molecules so far, the ones with the highest score. It speeds up the system's learning. The agent's remember policy is presented in equation 24 [24].

$$J(\theta) = \frac{(1-\alpha)}{N} \sum_{m \epsilon M} J_{mol}(\mathbb{A}, \mathbb{P}, A_m; \theta) + \frac{\alpha}{N} \sum_{\widetilde{m} \epsilon \widetilde{M}} J_{mol}(\mathbb{A}, \widetilde{\mathbb{A}}, \widetilde{A_m}; \theta) \quad (24)$$

Where, $J(\theta)$ is the agent's remember policy; $\alpha$ is a scaling factor based on the contribution of the best agent; $N$ is the number of molecules sampled; $M$ is the set of molecules, $m$, generated in the current agent's epoch; $J_{mol}$ is the policy for each molecule; $\mathbb{A}$ is the current agent; $\mathbb{P}$ is the previous model; $A_m$ is the set of actions chosen to build a molecule in the current agent's epoch; the "~" means that it is the best so far.

The molecule's remember policy is presented in equation 25 [24].

$$J_{mol}(\theta) = [\log P(\mathcal{B})_{\mathbb{B}} - (logP(\mathcal{B})_{Ref.} + \sigma \mathcal{S}(\mathcal{B}))]^2 \quad (25)$$

Where, $\sigma$ is a scaling factor treated as a hyperparameter, which tunes the contribution of the score; $P(\mathcal{B})_{\mathbb{B}}$ is the probability of choosing the sequence of actions $\mathcal{B}$ in the model $\mathbb{B}$; $P(\mathcal{B})_{Ref.}$ is the probability of the reference model for the same sequence of actions; $\mathcal{S}(\mathcal{B})$ is the score for the molecule generated by the actions $\mathcal{B}$.

The deep reinforcement learning will go through this loop until the defined number of epochs is completed. Every batch of reinforcement implies a new training of the generation model, utilizing the best model trained without reinforcement. The system's output, generative and reinforcement learning, is the best set of the requested number of new molecules to be designed.

The architecture of the deep reinforcement learning used is presented in Table 6.

*Table 1 - Deep Reinforcement Learning hyperparameters.*

| Parameters | Deep Reinforcement Learning's Value |
|---|---|
| A | 0.50 |
| Batch size | 20 |
| Block size | 1000 |
| Epochs | 500 |
| Generation epoch | 1040 |
| GGNN activation function | SELU |
| GGNN depth | 4 |
| GGNN dropout probability | 0 |
| GGNN hidden dimension | 250 |
| GGNN width | 100 |
| Initial learning rate | $1.00 \times 10^{-4}$ |
| Learning rate decay factor | 0.99 |
| Learning rate decay interval | 10 |
| Loss function | Kullback-Leibler divergence |
| Maximum relative learning rate | 1.00 |
| Message passing layers | 3 |
| Message size – input size of GRU | 100 |
| Minimum relative learning rate | $1.00 \times 10^{-4}$ |
| MLP activation function | SoftMax |
| MLP depth (Layers 1 and 2) | 4 |
| MLP dropout probability (Layers 1 and 2) | 0 |
| MLP hidden dimension (Layers 1 and 2) | 500 |
| Number of samples | 200 |
| Optimizer | Adam |
| $\sigma$ | 20 |
| Weight decay | 0 |
| Weight initialization | Uniform |

The reinforcement learning's epochs are subsequently to the generative ones in numeration. The neural network was trained for 500 epochs. Epoch 1040 was chosen as the agent model, given its performance in the evaluation step. The evaluation results for the selected epoch are presented in Table 2.

Table 2 –Evaluation results for the agent model chosen.

| Epoch | PV (0-1) | PVPT (0-1) | PPT (0-1) | $v_{av}$ | $\varepsilon_{av}$ | PU (0-1) |
|---|---|---|---|---|---|---|
| 1040 | 1.00 | 1.00 | 1.00 | 9.35 | 1.88 | 0.90 |

The training performance metric of learning rate are shown in Figure 2. It is possible to visualize that the chosen generation epoch has a high learning rate value, but is not the highest.

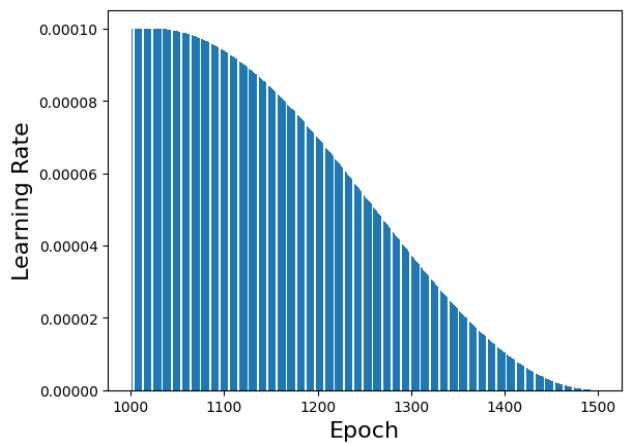

Figure 2 Reinforcement learning's learning rate.

Based on the best agent chosen, 200 molecules were generated, and 198 were considered valid by the network, Figures 3 to 6.

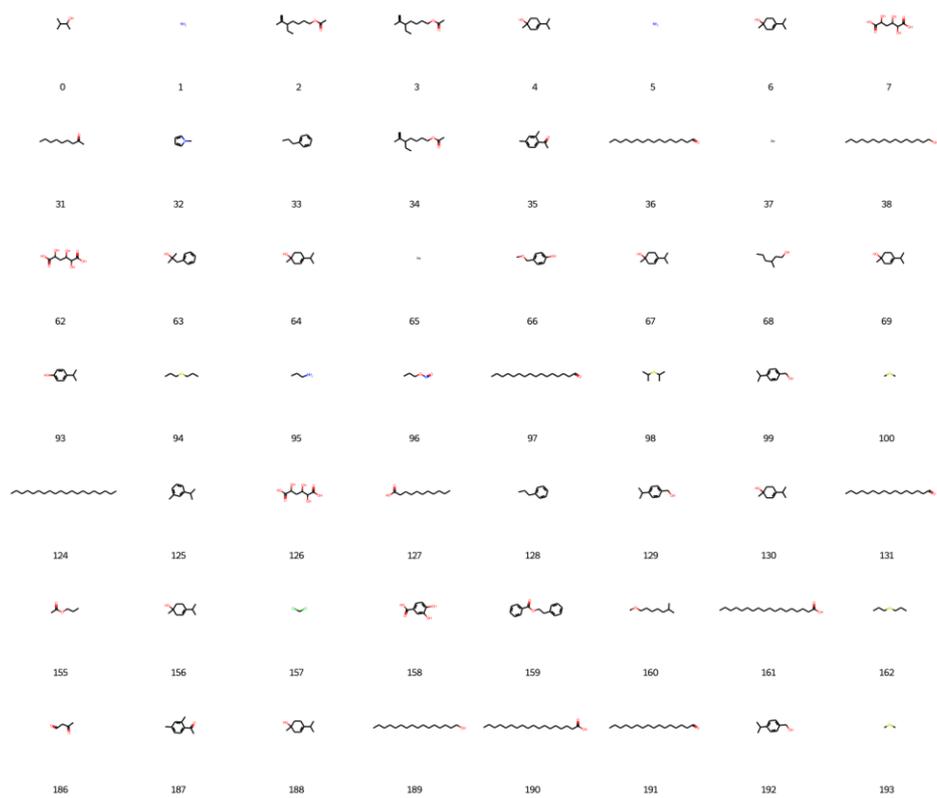

*Figure 3 New designed molecules from Deep Reinforcement Learning part 1.*

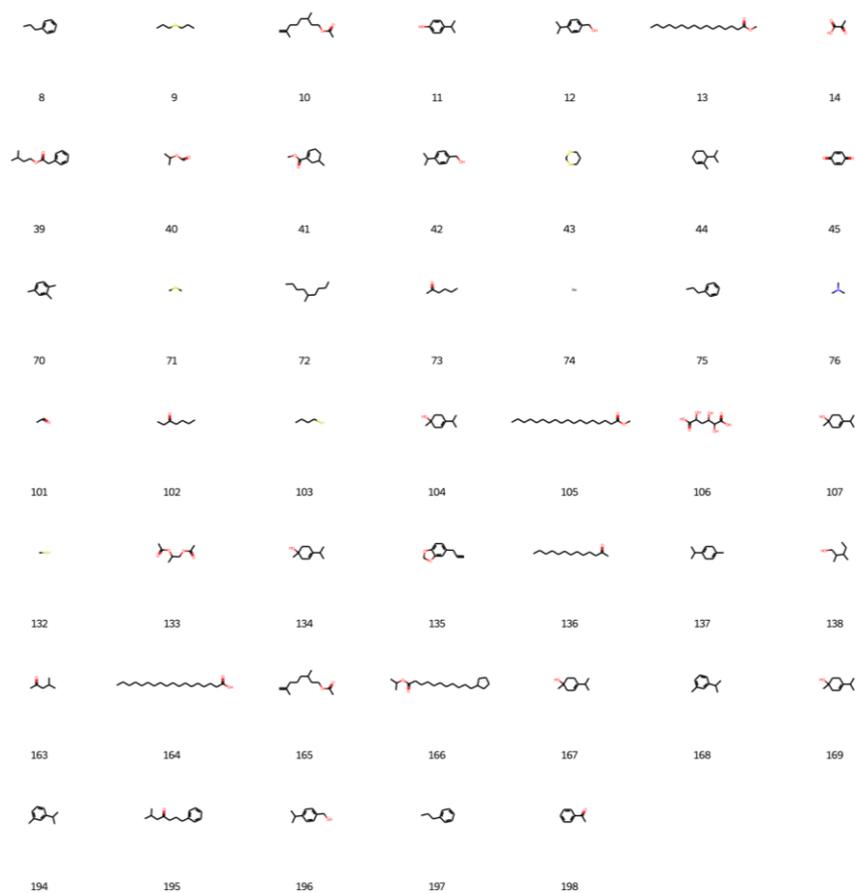

*Figure 4 New designed molecules from Deep Reinforcement Learning part 2.*

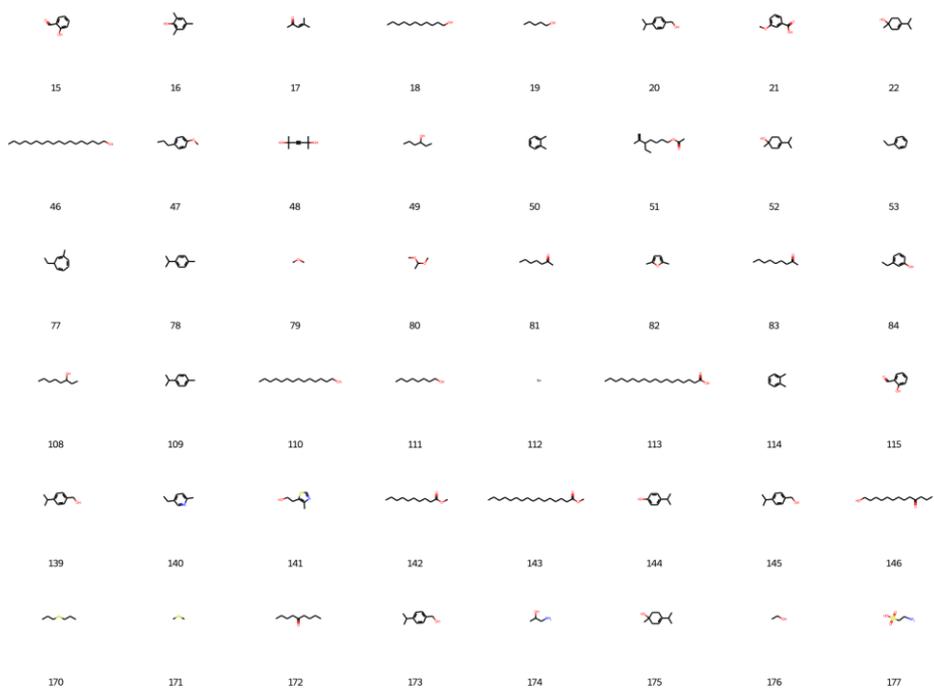

*Figure 5 New designed molecules from Deep Reinforcement Learning part 3.*

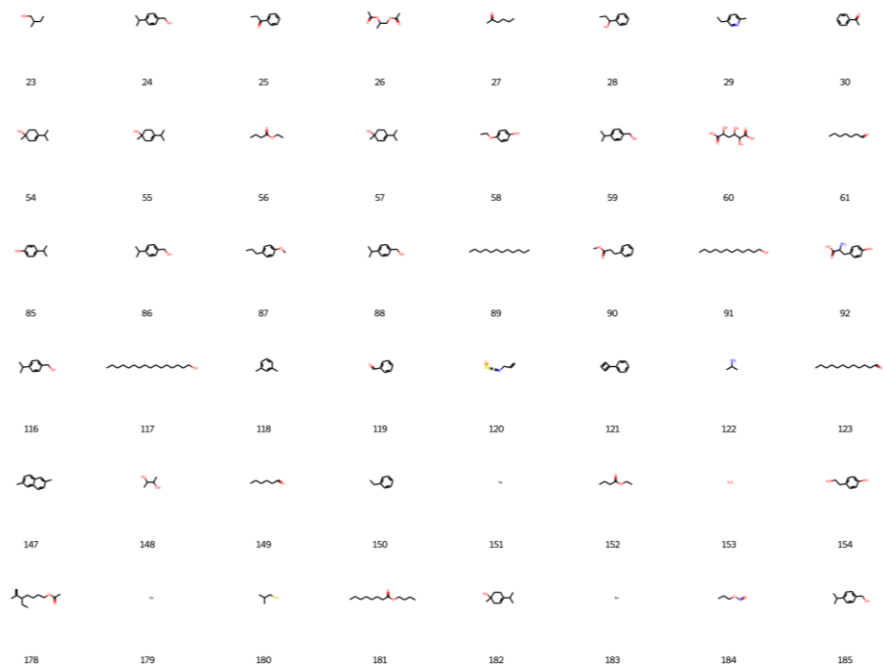

*Figure 6 New designed molecules from Deep Reinforcement Learning part 4.*

It is relevant to notice that the reinforcement learning training took approximately 6 hours and 18 minutes, while the deep generative model took about 4 days, 5 hours, and 31 minutes. Although the generative model epochs were double the reinforcement learning ones, the discrepancy in the amount of time required to train each SciML techniques is prominent.

As previously explained, during the reinforcement learning training, the molecules received a score based on the rewards metrics: SAScore, NPScore, and target size. The metrics SAScore and NPScore are the main results of this section, noticing that they are the target metrics of the work. The SAScore ranges from 1 to 10, being the lowest, the easiest to synthesize, and the highest value the hardest. The NPScore goes from -5 to 5; the higher the value more similar the newly designed molecule to a natural product. Based on the work of Chen et al. (2019) [59], a score higher than zero can be classified as a natural product. The pseudo-natural values are around zero on the scale with positive and negative values. Within this frame, values of SAScore below three and NPScore higher than zero will be considered optimal for the goal presented in this work, design molecules that are easy to be synthesized and that can be considered as natural.

To verify the impact of the reinforcement learning method on the desired results, the reward metrics were calculated for the molecules generated through the deep generative model in the last section. Figure 7 presents a heat map of the frequency of the SAScore range for generative and Reinforcement Learning (RL) results. In the same context, Figure 8 presents a heat map for the NPScore range.

In Figure 7, it is possible to visualize that 81.31 % of the molecules generated through the generative method obtained optimal values for the SAScore. In comparison, 79.80 % of the molecules obtained through reinforcement learning achieved this result. However, 57.58 % of the reinforcement learning's molecules are between 1 and 2 for the SAScore, while only 47.97 % of the generative method's molecules are in this range. This means that even though the generative model can design molecules that present good results in the target metric, the reinforcement learning design molecules with better performance.

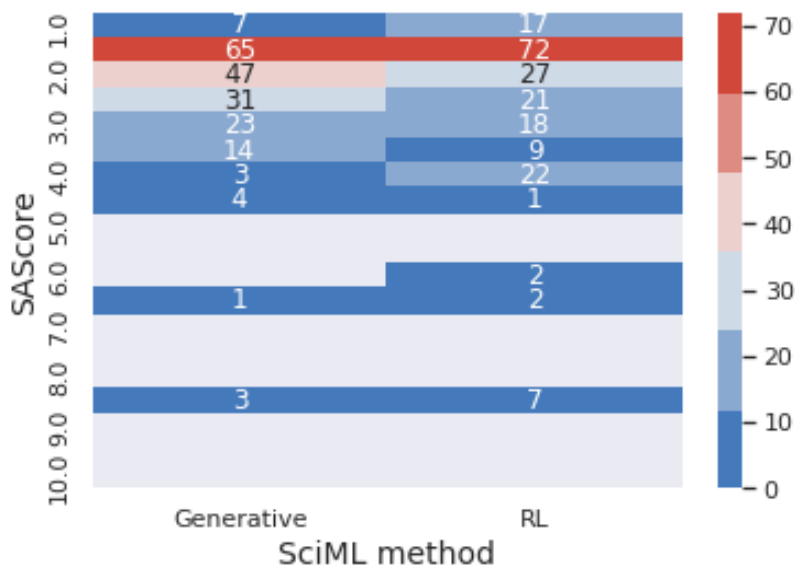

*Figure 7 Heatmap of the frequency of values in the range of the SAScore for the generative and the reinforcement learning model.*

In Figure 8, it is possible to conclude that, once again, the reinforcement learning designs molecules with better performance regarding the target metrics. The generative model presented 72.21 % of the molecules in the optimal range of results for the NPScore. However, 65.74 % of those are between 0 and 1, in the pseudo-natural range. At the same time, 74.75 % of the reinforcement learning molecules have optimal results, and only 58.80 % are in the pseudo-natural range. In achieving optimal results for both target metrics simultaneously, 53.53 % of the molecules achieved the optimal metrics for the generative, and 55.55 % achieved it for reinforcement learning. Demonstrating that using reinforcement learning improves the results when both metrics are considered. Furthermore, when reinforcement learning is used, there is a significantly higher number of molecules with the best values of the metrics. For instance, there are 20 molecules with NPscore of 3 against the 5 molecules provided by the solely generative approach.

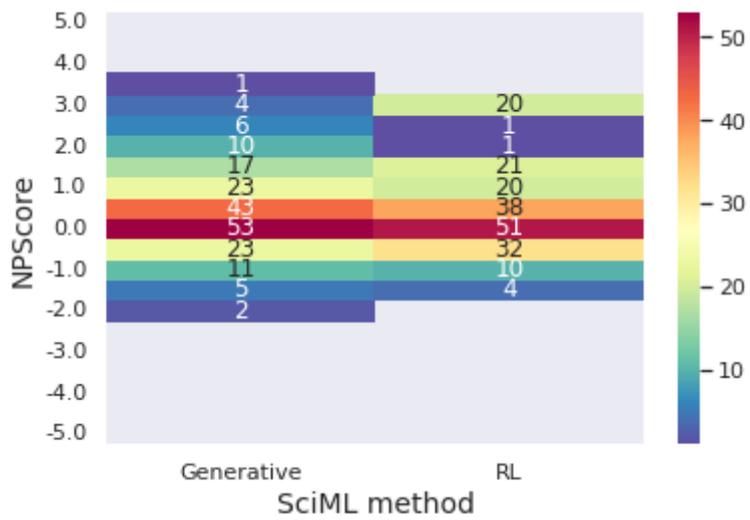

*Figure 8 Heatmap of the frequency of values in the range of the NPScore for the generative and the reinforcement learning model.*

## 3  Conclusion

The proposed work upraises a new perspective in flavor engineering based on Scientific Machine Learning. Combining deep generative models, reinforcement learning, and deep transfer learning was proposed to generate new flavor molecules. The focus was to generate new flavoring molecules that could be synthesized and applied in the natural products industry. Therefore, facing an emerging challenge present in the flavor engineering industry.

It was demonstrated that employing the proposed deep reinforcement learning method led to better results regarding the Synthetic Accessibility Score (SAScore) and the Natural Product-likeness Score (NPScore). It obtained 10 % more molecules within the optimal results than the deep generative model.

Hence, a new novel framework was created to enhance product engineering developments, providing an innovative approach to generating molecules with specific desired characteristics and chemical properties.

**Author Contributions:** Conceptualization, I.B.R.N., C.M.R, V.V.S, L.P.Q.; methodology, I.B.R.N., C.M.R., B.C.L.R., L.P.Q; writing—original draft preparation, I.B.R.N. and L.P.Q.; writing—review and ed-iting, I.B.R.N., L.P.Q. and A.M.R.; supervision, I.B.R.N., C.M.R., E.A.C., A.E.R., A.M.R. All au-thors have read and agreed to the published version of the manuscript.

**Funding:** This work was financially supported by: Project—NORTE-01-0145-FEDER-029384 funded by FEDER funds through NORTE 2020—Programa Operacional Regional do NORTE—and by national funds (PIDDAC) through FCT/MCTES. This work was also financially support-ed by: Base Funding—UIDB/50020/2020 of the Associate Laboratory LSRE-LCM—funded by na-tional funds through FCT/MCTES (PIDDAC), Capes for its financial support, financial code 001 and FCT—Fundação para a Ciência e Tecnologia under CEEC Institucional program.

**Conflicts of Interest:** The authors declare no conflict of interest.

# 4 References

1. T. Editors of Encyclopaedia Flavour. *Britannica* 2017.

2. Gary Reineccius *Flavor Chemistry and Technology*; Second Edi.; 2006; ISBN 9781566769334.

3. *Food Flavors Market Size, Share & COVID-19 Impact Analysis, By Type (Natural and Synthetic), By Application (Bakery, Beverages, Confectionery, Dairy, Convenience Food, Snacks, and Others), and Regional Forecast, 2021 - 2028*; 2022;

4. Sumesh Kumar, R.D. *Food Flavors Market by Type (Natural, and Artificial), and End-User (Beverages, Dairy & Frozen Products, Bakery & Confectionery, Savory & Snacks, Animal & Pet Food): Global Opportunity Analysis and Industry Forecast, 2021–2030*; 2021;

5. *Flavors & Fragrances Market by Ingredients (Natural, Synthetic), End Use (Beverage, Savory & Snacks, Bakery, Dairy Products, Confectionery, Consumer Products, Fine Fragrances), and Region (Asia Pacific, North America, Europe) - Global Forecast to 2026*; 2021;

6. Regulation (EC) No 1334/2008 of the European Parliament and of the Council of 16 December 2008 on Flavourings and Certain Food Ingredients with Flavouring Properties for Use in and on Foods and Amending Council Regulation (EEC) No 1601/91, Regulat. In; 2021; pp. 218–234.

7. Leong, Y.X.; Lee, Y.H.; Koh, C.S.L.; Phan-Quang, G.C.; Han, X.; Phang, I.Y.; Ling, X.Y. Surface-Enhanced Raman Scattering (SERS) Taster: A Machine-Learning-Driven Multireceptor Platform for Multiplex Profiling of Wine Flavors. *Nano Lett*. **2021**, *21*, 2642–2649, doi:10.1021/acs.nanolett.1c00416.

8. Bi, K.; Zhang, D.; Qiu, T.; Huang, Y. GC-MS Fingerprints Profiling Using Machine Learning Models for Food Flavor Prediction. *Processes* **2020**, *8*, 1–11, doi:10.3390/pr8010023.

9. Mousavi, S.S.; Schukat, M.; Howley, E. Deep Reinforcement Learning: An Overview. *Lect. Notes Networks Syst.* **2018**, *16*, 426–440, doi:10.1007/978-3-319-56991-8_32.

10. François-lavet, V.; Henderson, P.; Islam, R.; Bellemare, M.G.; François-lavet, V.; Pineau, J.; Bellemare, M.G. An Introduction to Deep Reinforcement Learning. (ArXiv:1811.12560v1 [Cs.LG]) Http://Arxiv.Org/Abs/1811.12560.

*Found. trends Mach. Learn.* **2018**, *II*, 1–140, doi:10.1561/2200000071.Vincent.

11. Yeh, J.F.; Su, P.H.; Huang, S.H.; Chiang, T.C. Snake Game AI: Movement Rating Functions and Evolutionary Algorithm-Based Optimization. *TAAI 2016 - 2016 Conf. Technol. Appl. Artif. Intell. Proc.* **2017**, 256–261, doi:10.1109/TAAI.2016.7880166.

12. Mnih, V.; Kavukcuoglu, K.; Silver, D.; Rusu, A.A.; Veness, J.; Bellemare, M.G.; Graves, A.; Riedmiller, M.; Fidjeland, A.K.; Ostrovski, G.; et al. Human-Level Control through Deep Reinforcement Learning. *Nature* **2015**, *518*, 529–533, doi:10.1038/nature14236.

13. Shao, K.; Tang, Z.; Zhu, Y.; Li, N.; Zhao, D. A Survey of Deep Reinforcement Learning in Video Games. **2019**, 1–13.

14. Zhou, Z.; Kearnes, S.; Li, L.; Zare, R.N.; Riley, P. Optimization of Molecules via Deep Reinforcement Learning. *Sci. Rep.* **2019**, *9*, 1–10, doi:10.1038/s41598-019-47148-x.

15. Dr. Ganesh Bagler FlavorDB Available online: https://cosylab.iiitd.edu.in/flavordb/ (accessed on 5 April 2022).

16. Li, Y.; Zemel, R.; Brockschmidt, M.; Tarlow, D. Gated Graph Sequence Neural Networks. *4th Int. Conf. Learn. Represent. ICLR 2016 - Conf. Track Proc.* **2016**, 1–20.

17. Pereira, T.; Abbasi, M.; Ribeiro, B.; Arrais, J.P. Diversity Oriented Deep Reinforcement Learning for Targeted Molecule Generation. *J. Cheminform.* **2021**, *13*, 1–17, doi:10.1186/s13321-021-00498-z.

18. Nwankpa, C.; Ijomah, W.; Gachagan, A.; Marshall, S. Activation Functions: Comparison of Trends in Practice and Research for Deep Learning. **2018**, 1–20.

19. Ertl, P.; Schuffenhauer, A. Estimation of Synthetic Accessibility Score of Drug-like Molecules Based on Molecular Complexity and Fragment Contributions. *J. Cheminform.* **2009**, *1*, 1–11, doi:10.1186/1758-2946-1-8.

20. Ertl, P.; Roggo, S.; Schuffenhauer, A. Natural Product-Likeness Score and Its Application for Prioritization of Compound Libraries. *J. Chem. Inf. Model.* **2008**, *48*, 68–74, doi:10.1021/ci700286x.

21. Chen, Y.; Stork, C.; Hirte, S.; Kirchmair, J. NP-Scout: Machine Learning Approach for the Quantification and Visualization of the Natural Product-Likeness of Small Molecules. *Biomolecules* **2019**, *9,* doi:10.3390/biom9020043.

22. Karageorgis, G.; Foley, D.J.; Laraia, L.; Waldmann, H. Principle and Design of Pseudo-Natural Products. *Nat. Chem.* **2020**, *12*, 227–235, doi:10.1038/s41557-019-0411-x.

23. M. D. Buhmann Radial Basis Functions. *Cambrity Univ. Press* **2001**, *Acta Numer*, 1–38, doi:10.1017/S0962492900000015.

24. Atance, S.R.; Diez, J.V.; Engkvist, O.; Olsson, S.; Mercado, R. De Novo Drug Design Using Reinforcement Learning with Graph-Based Deep Generative Models. *ChemRxiv* **2021**, 1–12.